\documentclass[10pt,twocolumn,letterpaper]{article}

\usepackage{cvpr}              

\usepackage{url}
\usepackage{graphicx}
\usepackage{booktabs}
\usepackage{colortbl}
\usepackage{pifont}
\usepackage{wrapfig}
\usepackage{multirow}
\usepackage{svg}
\usepackage[para]{footmisc}
\usepackage{amssymb}
\usepackage{longtable}

\definecolor{tablecolor}{HTML}{ccf2f5} 
\definecolor{tablecolor2}{HTML}{dcf5eb} 

\newcommand{\mymethod}{SAT}

\newcommand{\yes}{\color{blue}{\ding{51}}}
\newcommand{\no}{\color{red}{\ding{55}}}

\newcommand{\dd}[2]{$#1\scriptstyle{\pm#2}$}

\newcommand{\ddbfgreen}[2]{\cellcolor{tablecolor2}$\mathbf{#1\scriptstyle{\pm#2}}$}

\definecolor{cvprblue}{rgb}{0.21,0.49,0.74}
\usepackage[pagebackref,breaklinks,colorlinks,allcolors=cvprblue]{hyperref}

\def\paperID{30923} 
\def\confName{CVPR}
\def\confYear{2026}

\title{Structural Action Transformer for 3D Dexterous Manipulation}

\author{
    Xiaohan Lei$^1$ \quad
    Min Wang$^{2,}$\thanks{Corresponding authors.} \quad
    Bohong Weng$^1$ \quad
    Wengang Zhou$^{1,2,*}$ \quad
    Houqiang Li$^{1,2}$ \\[6pt]
    $^1$MoE Key Laboratory of Brain-inspired Intelligent Perception and Cognition, \\
    University of Science and Technology of China \\
    $^2$Institute of Artificial Intelligence, Hefei Comprehensive National Science Center \\[6pt]
    {\tt\small \{leixh, bhweng\}@mail.ustc.edu.cn} \quad
    {\tt\small wangmin@iai.ustc.edu.cn} \quad
    {\tt\small \{zhwg, lihq\}@ustc.edu.cn}\\[3pt]
    {\small Project Page: \quad \tt\small \url{https://xiaohanlei.github.io/projects/SAT}}
}

\begin{document}

\maketitle

\linespread{0.965}\selectfont

\begin{abstract}

Achieving human-level dexterity in robots via imitation learning from heterogeneous datasets is hindered by the challenge of cross-embodiment skill transfer, particularly for high-DoF robotic hands. 
Existing methods, often relying on 2D observations and temporal-centric action representation, struggle to capture 3D spatial relations and fail to handle embodiment heterogeneity.
This paper proposes the Structural Action Transformer (SAT), a new 3D dexterous manipulation policy that challenges this paradigm by introducing a structural-centric perspective. 
We reframe each action chunk not as a temporal sequence, but as a variable-length, unordered sequence of joint-wise trajectories.
This structural formulation allows a Transformer to natively handle heterogeneous embodiments, treating the joint count as a variable sequence length.
To encode structural priors and resolve ambiguity, we introduce an Embodied Joint Codebook that embeds each joint's functional role and kinematic properties.
Our model learns to generate these trajectories from 3D point clouds via a continuous-time flow matching objective.
We validate our approach by pre-training on large-scale heterogeneous datasets and fine-tuning on simulation and real-world dexterous manipulation tasks.
Our method consistently outperforms all baselines, demonstrating superior sample efficiency and effective cross-embodiment skill transfer.
This structural-centric representation offers a new path toward scaling policies for high-DoF, heterogeneous manipulators.

\end{abstract}

\begin{figure}[t]
  \centering
  \includegraphics[width=0.99\linewidth]{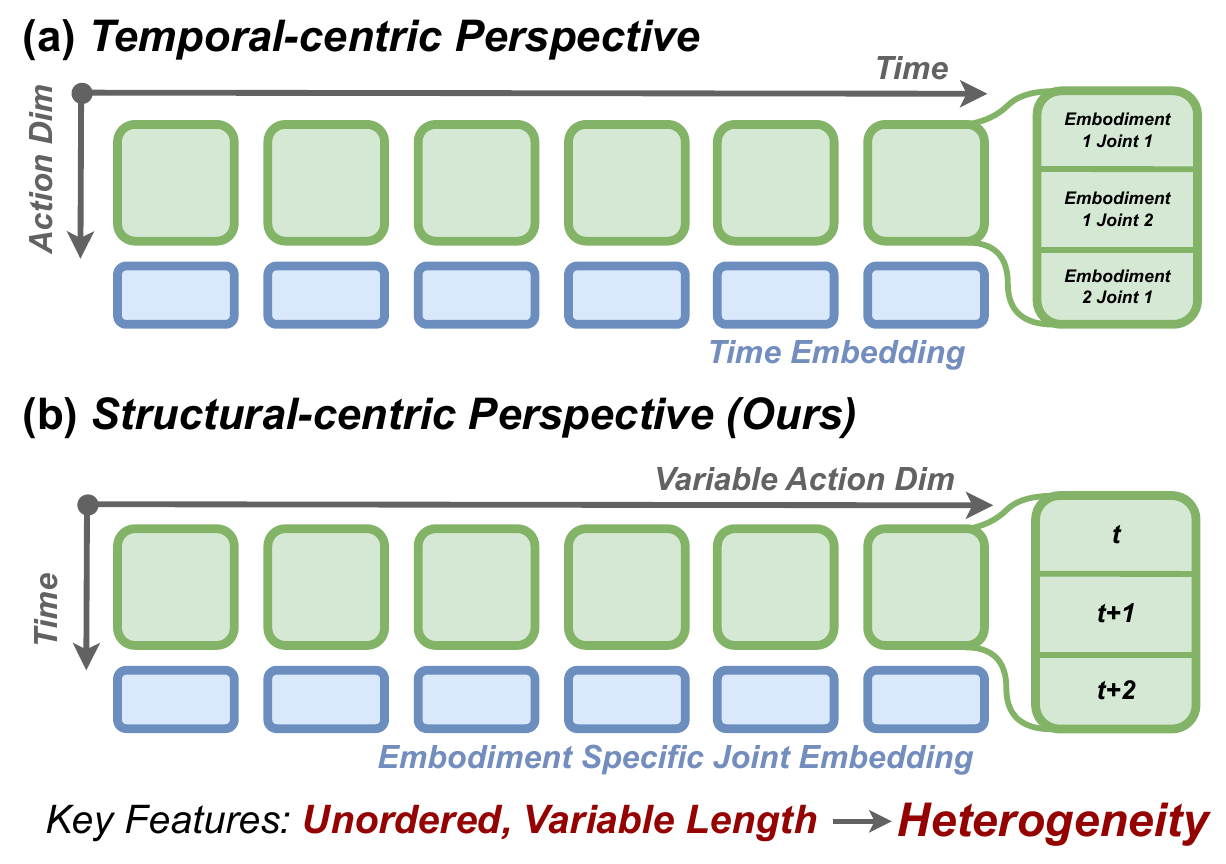}
  \vspace{-8pt}
  \caption{Conceptual illustration of action chunk tokenization. 
\textbf{(a)} The conventional temporal-centric perspective, which structures actions as a sequence of $T$ timesteps (chunk length), with each token having dimension $D_a$ (action dim). 
\textbf{(b)} Our proposed structural-centric perspective, which reframes the action chunk as a sequence of $D_a$ joints, where each token's feature is its temporal trajectory over $T$. This $(D_a, T)$ view naturally handles heterogeneous embodiments as a variable-length, unordered sequence, which is a key feature of our approach.}
  \vspace{-26pt}
  \label{fig:teaser}
\end{figure}    
\section{Introduction}

The quest for robotic systems capable of human-level dexterity represents a grand challenge in Embodied Artificial Intelligence. Dexterous robotic hands, with their high degrees of freedom (DoF), hold the potential to perform a vast array of complex, contact-rich manipulation tasks currently beyond the reach of simpler parallel-jaw grippers. Imitation learning, particularly in its offline formulation, has emerged as a promising paradigm for teaching such skills by leveraging large-scale datasets of human demonstrations~\cite{o2024open, team2024octo, khazatsky2024droid, dasari2019robonet, walke2023bridgedata, mandlekar2018roboturk, fang2023rh20t, jiang2025kaiwu, bu2025agibot}. A critical bottleneck, however, lies in the cross-embodiment transfer of these skills: how can a robot effectively learn from heterogeneous demonstrations, given the significant differences in morphology, kinematics, and sensory feedback~\cite{wang2024scaling, zhou2025mitigating, chen2025vidbot, yang2025egovla}? While recent Vision-Language-Action (VLA) models have made strides in general-purpose robotics~\cite{zitkovich2023rt, zheng2024tracevla, kim2024openvla, wen2025tinyvla, zhao2025cot}, they predominantly rely on 2D visual inputs, which often fail to capture the intricate 3D spatial relationships essential for precise dexterous manipulation~\cite{qian20253d, li2025bridgevla, li2025pointvla, huang2023voxposer, zhen20243d}. This work tackles the challenge of cross-embodiment imitation for dexterous hands by directly learning from 3D point cloud observations, proposing a fundamental shift in how we represent and process robotic actions.

A dominant paradigm in recent policy learning is action chunking, where a model predicts a sequence of future actions $(T, D_a)$~\cite{chi2023diffusion, zhao2023learning}. As in Figure~\ref{fig:teaser} (a), this \textbf{temporal-centric perspective}, which treats each $D_a$-dimensional vector as a token in a temporal sequence, is effective for low-dimensional systems. However, this representation faces a fundamental challenge when scaling to high-DoF, heterogeneous manipulators. As the action dimension $D_a$ grows (\textit{e.g.}, from a 7-DoF robot arm to a 24-DoF dexterous hand), the model must learn complex, implicit correlations within a monolithic feature vector. More critically, this fixed-dimensional view provides no natural mechanism for cross-embodiment transfer, as different morphologies cannot be directly compared. 

This paper challenges this conventional wisdom by reframing the action representation problem. We propose a \textbf{structural-centric perspective}, modeling actions as a sequence of $D_a$ embodied joints, where each joint's feature is its trajectory over the time horizon $T$, \textit{i.e.}, $(D_a, T)$, as shown in Figure~\ref{fig:teaser} (b). This structural view directly addresses the heterogeneity problem: the sequence length $D_a$ can now vary between embodiments, a property that Transformer architectures handle natively~\cite{vaswani2017attention}. This allows a policy to learn transferable skills by finding functional similarities between corresponding joints. At the same time, this formulation also allows the model to learn a compressed representation of motion primitives for each joint, as time is now treated as the feature dimension.

We introduce the Structural Action Transformer (SAT), a novel 3D dexterous manipulation policy built upon this structural-centric representation.
Our model takes raw 3D point cloud observations and natural language instructions as input.
The 3D scene is processed by a hierarchical tokenizer that uses Farthest Point Sampling (FPS) and PointNets~\cite{qi2017pointnet} to extract both local geometric tokens and a single global scene token.
These are combined with language features from a T5 encoder~\cite{raffel2020exploring} to form a multi-modal observation sequence.
This sequence conditions a Diffusion Transformer (DiT)~\cite{peebles2023scalable} that operates directly on our $(D_a, T)$ structural action representation.
We treat the $D_a$ joints as a variable-length sequence, where each token represents a compressed temporal trajectory for a single joint.
To explicitly encode structural priors and manage heterogeneity, we introduce an Embodied Joint Codebook.
This codebook provides learnable embeddings for each joint based on its morphological properties, enabling the model to identify functional correspondences across different embodiments.
The policy is trained to generate the entire action chunk by learning a conditional velocity field via a continuous-time flow matching objective~\cite{lipman2022flow}, with the final action produced by an ODE solver.

We demonstrate the efficacy of this approach by pre-training on large-scale, heterogeneous datasets of both human and robot demonstrations~\cite{liu2022hoi4d, grauman2024ego, pan2023aria, fourier2025actionnet, wang2024dexcap} and then fine-tuning it on a suite of challenging simulation benchmarks~\cite{rajeswaran2017learning, bao2023dexart, chen2023bi} and real-world bimanual manipulation tasks.
Our extensive experiments show that this structural-centric paradigm consistently outperforms strong baselines~\cite{chi2023diffusion, wang2024scaling, zheng2025universal, ze20243d, yan2025maniflow} and achieves superior sample efficiency, proving its effectiveness in cross-embodiment skill transfer.
Fundamentally, our work is the first to successfully implement a policy that tokenizes actions along the structural dimension, offering a new and scalable path toward learning generalist policies for a diverse ecosystem of high-DoF, heterogeneous manipulators.

\vspace{-4pt}
\section{Related Work}
\label{sec:related_work}


\subsection{Action Representation in Policy Learning}

The representation of actions is a fundamental design choice in learning-based control. While early methods in imitation learning predict a single action at each timestep~\cite{pomerleau1988alvinn, ross2011reduction}, this autoregressive approach is known to suffer from compounding errors, where small inaccuracies accumulate over long horizons, leading to significant trajectory divergence~\cite{ross2010efficient}.

To mitigate this issue, a dominant paradigm in recent literature is \textbf{action chunking}, where a policy outputs a sequence of future actions at each inference step~\cite{sharma2019dynamics, zhao2023learning, chi2023diffusion}. This approach has been shown to improve temporal consistency and reduce the impact of compounding errors. For instance, Diffusion Policy~\cite{chi2023diffusion} and related diffusion-based models~\cite{janner2022planning, reuss2023goal, wen2024diffusion, hou2025dita, liu2024rdt, liu2025hybridvla} generate chunked action trajectories by iteratively denoising a random tensor, producing smooth and effective behaviors. Similarly, transformer-based models~\cite{zhao2023learning, reed2022generalist, brohan2022rt, kim2024openvla, zitkovich2023rt} predict sequences of discretized action tokens. This temporal-centric view, structuring action chunks as a sequence of feature vectors over time, $(T, D_a)$, is now standard in large-scale robotic policies.

While effective, these methods treat the action vector at each timestep as a monolithic entity. This overlooks the rich kinematic structure of the robot and becomes inefficient as the action dimensionality $D_a$ grows. Our work challenges this conventional representation. Instead of viewing actions as a sequence of temporal snapshots, we propose to model them as a sequence of joint-wise trajectories, $(D_a, T)$. This reframing allows the model to learn compressed temporal primitives for each joint, and more importantly, provides a natural mechanism to handle embodiment heterogeneity.

\subsection{Dexterous Manipulation}

Dexterous robotic hands, with their high degrees of freedom, offer the potential to replicate human-level dexterity but pose significant learning challenges due to their complex dynamics and contact-rich nature \cite{bicchi2000robotic}. Early successes often relied on precise analytical models \cite{murray2017mathematical}, but recent progress has been dominated by learning-based methods.

Deep reinforcement learning (RL) has been successfully used to learn complex in-hand manipulation skills from scratch in simulation \cite{andrychowicz2020learning, rajeswaran2017learning, nagabandi2020deep, luo2025precise}, though transferring these policies to the real world remains a challenge. Imitation learning from human demonstrations offers a more data-efficient alternative. This includes learning from various interfaces such as vision-based motion capture \cite{qin2023anyteleop, cheng2024open, wang2024dexcap, yang2025ace}, or glove-based controllers~\cite{zhang2025doglove, gao2025glovity, xu2025dexumi}. 

Recent Vision-Language-Action (VLA) models have also been applied to this domain~\cite{yang2025egovla, wen2025dexvla, jang2025dreamgen, bjorck2025gr00t}. For example, DexGraspVLA \cite{zhong2025dexgraspvla} and DexVLG~\cite{he2025dexvlg} leverage diffusion-based or flow-based policies and large pre-trained models to achieve zero-shot grasping of novel objects. 
However, nearly all of these methods still rely on
the conventional $(T, D_a)$ action representation. This approach treats the action as a monolithic, fixed-dimensional vector at each timestep, which is fundamentally ill-suited for cross-embodiment transfer as it provides no natural mechanism to align or compare manipulators with different kinematic structures or joint counts.

\subsection{Heterogeneous Learning}

A central goal of modern robotics is to create ``generalist'' agents that can operate across a wide range of tasks, environments, and robot morphologies \cite{bommasani2021opportunities}. This has led to a paradigm shift towards pre-training large policies on massive, heterogeneous datasets. The Open X-Embodiment dataset \cite{o2024open} represents a landmark effort in this direction, aggregating data from dozens of different robots.

Several strategies have emerged to handle the ``heterogeneity'' problem. Data-centric approaches aim to unify diverse datasets under a common data format and action representation. For instance, some methods~\cite{brohan2022rt, o2024open, kim2024openvla, pertsch2025fast, lee2024behavior, wang2025vq} discretize the action space into a shared vocabulary of tokens , while others~\cite{zitkovich2023rt, driess2023palm, zhou2025physvlm, li2024manipllm, han2024dual, team2025gemini, ji2025robobrain, nair2022r3m} leverage the power of pre-trained vision-language models (VLMs) to perform high-level reasoning, effectively treating robot control as a sequence modeling problem. This paradigm has been successfully scaled by training single, diffusion-based generalist policies on large-scale aggregated datasets~\cite{team2024octo, black2410pi0, intelligence2025pi}. Modular approaches explicitly design architectures to handle heterogeneity. A common technique is to introduce embodiment-specific ``stems'' that tokenize proprioceptive and visual inputs from different robots into a shared latent space, which is then processed by a large, pre-trained ``trunk''~\cite{wang2024scaling, zheng2025universal}

While these methods have achieved impressive generalization, they either enforce a unified action space that may not be optimal for all robots, or rely on separate modules to align inputs before the core policy network. Our work introduces a fundamentally different approach. By re-structuring the action prediction problem as generating a sequence of joint trajectories, $(D_a, T)$, we leverage a core property of the Transformer architecture: its ability to operate on variable-length sequences. In our framework, a robot's embodiment is defined by the length of the sequence, $D_a$. This allows a single, unified policy to naturally handle different robots and enables the self-attention mechanism to learn functional similarities and mappings between the joints of different embodiments directly within the representation space.
\vspace{-4pt}
\section{Method}
\label{sec:method}

\begin{figure*}[t]
  \centering
  \includegraphics[width=0.99\linewidth]{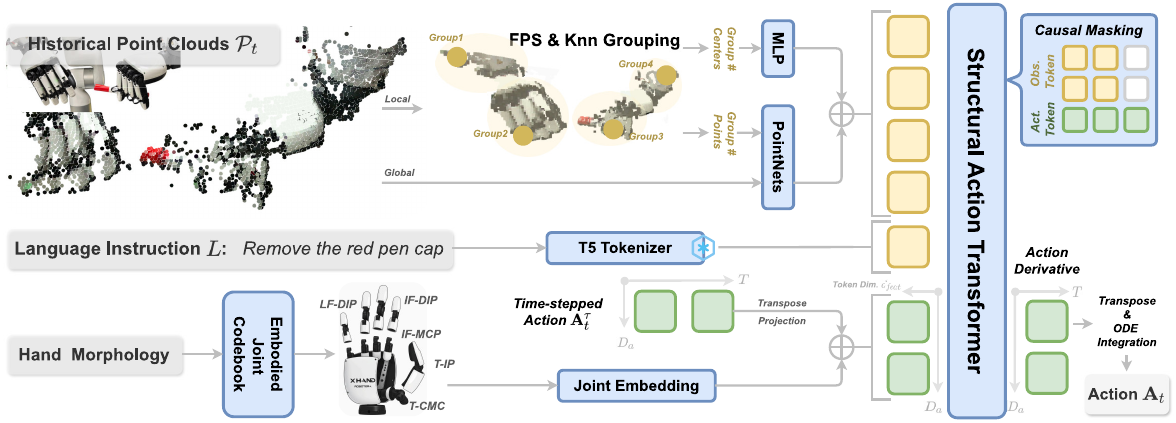}
  \vspace{-8pt}
  \caption{Our proposed model architecture. The policy takes a history of $T_o$ raw 3D point clouds $\mathcal{P}_t = (\mathbf{P}_{t-T_o+1}, \dots, \mathbf{P}_t)$ and a language instruction $L$ as input. 
 \textit{Observation Tokenizer}: Each point cloud $\mathbf{P}_k$ in the history is processed via Farthest Point Sampling (FPS) and PointNets to extract local geometric tokens and a global scene context. The tokens from each time step are concatenated to form the final observation token sequence.
 Language is encoded by a T5 tokenizer~\cite{2020t5}. 
 \textit{Structural Action Tokenizer}: 
 Guided by the manipulator’s morphology, the Embodied Joint Codebook produces structural-centric embeddings aligned with the action dimension $D_a$, which are added to the time-stepped noisy tokens $\mathbf{A}_t^\tau$.
 \textit{Structural Action Transformer:} A
 DiT~\cite{peebles2023scalable} with causal masking predicts the action velocity
 field. This field is then integrated via an ODE solver to produce the final
 action chunk $\mathbf{A}_t$.}
  \vspace{-16pt}
  \label{fig:main}
\end{figure*}


\subsection{Problem Formulation}

We formulate the control problem for
dexterous hands as a conditional generative modeling task. At each time step
$t$, the policy receives an observation $o_t$ which consists of a history of the last $T_o$ raw 3D point clouds, $\mathcal{P}_t = (\mathbf{P}_{t-T_o+1}, \dots, \mathbf{P}_t)$, where each $\mathbf{P}_k \in \mathbb{R}^{N \times 3}$ ($N$ is the
number of points), and a natural language instruction $L$. The goal is to learn a policy $\pi$ that
predicts a chunk of future actions $\mathbf{A}_t$. 
As motivated in our
introduction, we challenge the conventional temporal-centric action
representation $(T, D_a)$. Instead, we define the action chunk from a structural-centric
perspective as $\mathbf{A}_t \in \mathbb{R}^{D_a \times T}$, where $D_a$ is the
dimensionality of the robot's action space and $T$ is the prediction horizon.
Each row $j_i \in \mathbb{R}^{T}$ of $\mathbf{A}_t$ represents the entire
future trajectory for the $i$-th joint. The policy $\pi$ thus models the
conditional distribution 
$p(\mathbf{A}_t | o_t)$. 
During closed-loop control, this action chunk $\mathbf{A}_t$
is generated in a receding-horizon manner. A subset of the predicted actions is
executed, and the policy is re-queried with the new observation, enabling
continuous feedback and correction.

\subsection{Policy as a Conditional Normalizing Flow}

We model the complex, high-dimensional conditional distribution $p(\mathbf{A}_t | o_t)$ using a continuous-time normalizing flow (CNF)~\cite{lipman2022flow}. We frame this as learning a conditional velocity field $v(\mathbf{A}_t^\tau, \tau, o_t)$ that transports a standard Gaussian $\mathcal{N}(0, I)$ to the action distribution. We parameterize the velocity field with a neural network $\epsilon_\theta(\mathbf{A}_t^\tau, \tau, o_t)$, where $\tau \in [0, 1]$ is the flow time. The network is trained to minimize the following objective:
\begingroup
\setlength{\abovedisplayskip}{-0pt}
\setlength{\belowdisplayskip}{2pt}
\begin{equation}\label{eq:fm_objective}
\resizebox{\linewidth}{!}{$
\mathcal{L}(\theta)=\mathbb{E}_{\tau\sim\mathcal{U}(0,1),\mathbf{A}_t^0\sim\mathcal{N}(0,I),\mathbf{A}_t^1\sim\mathcal{D}}
\left[\left\|\epsilon_\theta(\mathbf{A}_t^\tau,\tau,o_t)-(\mathbf{A}_t^1-\mathbf{A}_t^0)\right\|^2\right],
$}
\end{equation}
\endgroup
where $\mathcal{D}$ is the dataset of ground-truth action chunks, $\mathbf{A}_t^1 \sim \mathcal{D}$ is a ground-truth action, $\mathbf{A}_t^0 \sim \mathcal{N}(0,I)$ is a noise sample, and $\mathbf{A}_t^\tau = (1-\tau)\mathbf{A}_t^0 + \tau \mathbf{A}_t^1$. The conditioning observation $o_t$ is the one associated with the target action $\mathbf{A}_t^1$.

At inference time, we sample a noise $\mathbf{A}_t^0 \sim \mathcal{N}(0,I)$ and generate the final action chunk $\mathbf{A}_t^1$ by solving the ordinary differential equation (ODE) $\frac{d\mathbf{A}_t^\tau}{d\tau} = \epsilon_\theta(\mathbf{A}_t^\tau, \tau, o_t)$ from $\tau=0$ to $\tau=1$. This is done with a numerical solver, such as a single Euler integration step, which recovers the probability flow in one function evaluation (1-NFE).

\subsection{Policy Network Architecture}

As shown in Figure~\ref{fig:main}, we design our velocity field model
$\epsilon_{\theta}$ as a Transformer~\cite{vaswani2017attention} architected to
ingest our novel structural action tokens and the multi-modal conditioning
observation tokens. The architecture is divided into three components: an
\textbf{Observation Tokenizer} that tokenizes the historical 3D point clouds
and language inputs, 
a \textbf{Structural Action Tokenizer} that embed structural priors into action tokens,
and a \textbf{Structural Action Transformer} that predicts the
action velocity field conditioned on the observations.

\vspace{-2pt}
\subsubsection{Observation Tokenizer}
\vspace{-2pt}

Our hierarchical point cloud tokenizer is designed to
capture both local geometric details and the global scene context from the entire observation history $\mathcal{P}_t$. We apply a shared encoding module to each of the $T_o$ point clouds $\mathbf{P}_k \in \mathcal{P}_t$. For a given $\mathbf{P}_k$, we first use
Farthest Point Sampling (FPS) to select $M$ local group centers $C_k = \{c_1,
..., c_M\} \subset \mathbf{P}_k$. For each center $c_{i}$, we query its $K$
nearest neighboring points to form a local group $\mathbf{P}_{k,i} \subset
\mathbf{P}_k$. Each local group $\mathbf{P}_{k,i}$ is processed by a shared
PointNet~\cite{qi2017pointnet} to extract a local geometric feature
$f_{k,i}\in\mathbb{R}^{d_{feat}/T_o}$. The 3D coordinates of the group centers $C_k$
are passed through an MLP to create positional embeddings
$p_{k,i}\in\mathbb{R}^{d_{feat}/T_o}$. These are combined
($f_{k,i}^{\prime}=f_{k,i}+p_{k,i}$) as $M$ local point tokens,
$tok_{l,k}\in\mathbb{R}^{M\times d_{feat}/T_o}$. To leverage the
permutation-invariant nature of these local patches, we apply a random shuffle
to the $tok_{l,k}$ sequence during training as data augmentation. In parallel,
to capture holistic scene understanding, we feed the entire raw point cloud
$\mathbf{P}_k$ into a separate PointNet-based encoder~\cite{qi2017pointnet}.
This module produces a single global token, $tok_{g,k} \in
\mathbb{R}^{d_{feat}/T_o}$, which represents the overall scene context for that timestep. This global
token does not receive any positional embedding.

This process yields $T_o$ global tokens and $T_o$ sets of $M$ local tokens. These are concatenated to form the full point cloud history sequence: 
\begingroup
\setlength{\abovedisplayskip}{1pt}
\setlength{\belowdisplayskip}{2pt}
\begin{equation}\label{eq:hist_token}
\begin{aligned}
tok&_{hist} = \text{Cat}\big(\text{Cat}(tok_{g,\,t-T_o+1},\dots,tok_{g,\,t}),\\
&\text{Cat}(tok_{l,\,t-T_o+1},\dots,tok_{l,\,t})\big)
\in\mathbb{R}^{(1+M)\times d_{feat}},
\end{aligned}
\end{equation}
\endgroup
where Cat means concatenation.
The language instruction $L$ is
tokenized and encoded using a pre-trained T5 encoder~\cite{2020t5} to produce a
sequence of language tokens $tok_{lang}\in\mathbb{R}^{L_{lang}\times
d_{feat}}$. The final observation tokens are formed by concatenating the
historical tokens and the language tokens: $tok_{obs} =
\text{Cat}(tok_{hist}, tok_{lang})$. This sequence serves
as the conditioning prefix for the model.

\vspace{-2pt}
\subsubsection{Structural Action Tokenizer}
\vspace{-2pt}

The noisy action $\mathbf{A}_{t}^\tau
\in\mathbb{R}^{D_{a}\times T}$ is treated as a sequence of $D_{a}$ tokens,
where each token represents the entire temporal trajectory for a single joint.
This high-dimensional temporal vector (dim $T$) contains significant
redundancy. We compress it by passing each of the $D_a$ joint trajectories
through a shared MLP, projecting it from $T$ dimensions to a lower-dimensional
embedding $d_{feat}$ (\textit{e.g.}, from $64$ to $16$). This results in an
embedded action sequence $tok_{act}\in \mathbb{R}^{D_{a}\times d_{feat}}$.

To resolve the ambiguity between the $D_a$
joint tokens and explicitly embed structural priors, we introduce an
\textbf{Embodied Joint Codebook}. This codebook is derived from the
manipulator's morphology. For any given hand, we define each joint $j$ as a
three-part triplet $J_{j} = (e, f, r)$:
\begin{itemize}
   \item $e\in\mathbb{Z}$: The \textbf{Embodiment ID}, a unique identifier
for the manipulator (\textit{e.g.}, ShadowHand, XHand).
   \item $f\in\mathbb{Z}$: The \textbf{Functional Category}. Inspired by
human hand anatomy, we classify joints by their functional role, such as
Carpometacarpal (CMC), Metacarpophalangeal (MCP), Proximal Interphalangeal
(PIP), or Distal Interphalangeal (DIP) joints.
   \item $r\in\mathbb{Z}$: The \textbf{Rotation Axis}, describing the
joint's primary motion, \textit{e.g.}, Flexion/Extension, Abduction/Adduction,
or Pronation/Supination.
\end{itemize}
The complete mapping table covering embodiment, function category and rotation axis used in our experiments is provided in Appendix.
Each element in the triplet $(e,f,r)$
indexes a separate learnable embedding table. The final codebook embedding $C_{j}\in\mathbb{R}^{d_{feat}}$ for
joint $j$, is the sum of its three component
embeddings. This design is central to handling heterogeneity: two different
hands (different $e$) may share the same functional joint (same $f$) and
rotation (same $r$), resulting in similar codebook embeddings that prime the
model for transfer learning. The final input sequence for the action
tokens is formed by adding the codebook embeddings to the compressed action
trajectories: $tok_{input\_act} = tok_{act} + \mathbf{E}$, where
$\mathbf{E} \in \mathbb{R}^{D_a \times d_{feat}}$ is the matrix of codebook
embeddings for the manipulator. 

\vspace{-2pt}
\subsubsection{Structural Action Transformer}
\vspace{-2pt}

The $tok_{input\_act}$ sequence is then
concatenated with the observation tokens $tok_{obs}$. This combined sequence
is fed into the DiT~\cite{peebles2023scalable}. 
We modify the DiT's
self-attention mask to enforce causal masking, where $tok_{obs}$ only
attends to other observation tokens, 
and $tok_{input\_act}$ attends to
all observation tokens and all other action tokens. 
The output tokens from the
DiT corresponding to the action sequence are passed through a final MLP to produce
the predicted action velocity field
$\epsilon_{\theta}(\mathbf{A}_{t}^{\tau},\tau,o_{t})$.

\begin{table*}[t]
\centering
\small
\resizebox{0.99\textwidth}{!}{
\begin{tabular}{l|c|c|c|c|c|c}
\toprule
  Method & Params (M) & Modality & Adroit (3)~\cite{rajeswaran2017learning} & DexArt (4)~\cite{bao2023dexart} & Bi-DexHands (4)~\cite{chen2023bi} & Average Success  \\
\midrule
  Diffusion Policy~\cite{chi2023diffusion} & 266.8 & 2D & \dd{0.32}{0.03} & \dd{0.49}{0.04} & \dd{0.42}{0.05} & \dd{0.42}{0.04} \\
  HPT~\cite{wang2024scaling} & 13.99 & 2D & \dd{0.45}{0.02}  & \dd{0.53}{0.05} & \dd{0.44}{0.04} & \dd{0.47}{0.04}  \\
  UniAct~\cite{zheng2025universal} & 1053 & 2D & \dd{0.49}{0.01} & \dd{0.55}{0.03} & \dd{0.47}{0.07} & \dd{0.50}{0.05} \\
  
  3D Diffusion Policy~\cite{ze20243d} & 255.2 & 3D & \dd{0.68}{0.03} & \dd{0.69}{0.02} & \dd{0.55}{0.14} & \dd{0.63}{0.06} \\
  3D ManiFlow Policy~\cite{yan2025maniflow} & 218.9 & 3D & \dd{0.70}{0.02} & \dd{0.70}{0.03} & \dd{0.59}{0.07} & \dd{0.66}{0.04} \\
  \textbf{\mymethod{} (Ours)} & 19.36 & 3D & \ddbfgreen{0.75}{0.02} & \ddbfgreen{0.73}{0.03} & \ddbfgreen{0.67}{0.05} & \ddbfgreen{0.71}{0.04} \\
\bottomrule


\end{tabular}
}
\vspace{-8pt}
\caption{Quantitative comparison of our method against 2D (image-based) and 3D (point cloud-based) baselines on 11 dexterous manipulation tasks from the Adroit~\cite{rajeswaran2017learning}, DexArt~\cite{bao2023dexart}, and Bi-DexHands~\cite{chen2023bi} simulation benchmarks.}
\vspace{-16pt}
\label{tab:sim}
\end{table*}

\vspace{-4pt}
\section{Experiments}
\vspace{-2pt}
\label{sec:experiments}

\subsection{Datasets and Experimental Setup}

\begin{figure}[t]
  \centering
  \includegraphics[width=0.99\linewidth]{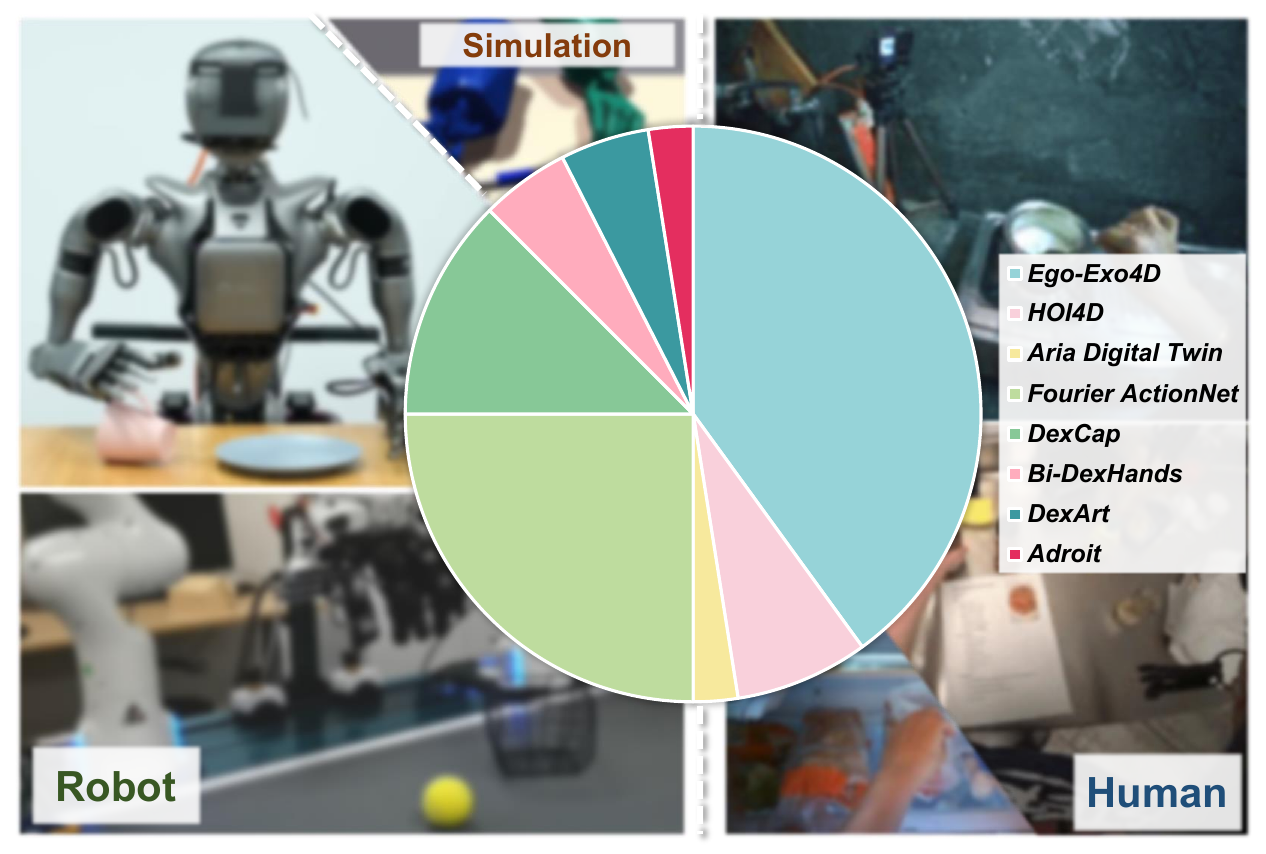}
  \vspace{-8pt}
  \caption{Composition of the offline pre-training dataset. The pie chart illustrates the relative data scale of each of the constituent datasets~\cite{liu2022hoi4d, grauman2024ego, pan2023aria, fourier2025actionnet, wang2024dexcap, rajeswaran2017learning, bao2023dexart, chen2023bi}.}
  \vspace{-16pt}
  \label{fig:dataset_count}
\end{figure}

\noindent\textbf{Offline Pre-training Datasets.}
Our approach leverages large-scale, heterogeneous datasets for pre-training. 
The full dataset is a mixture from three distinct sources, as detailed in Figure~\ref{fig:dataset_count}. For \textbf{human demonstrations}, we use HOI4D~\cite{liu2022hoi4d}, Ego-Exo4D~\cite{grauman2024ego} and Aria Digital Twin (ADT)~\cite{pan2023aria}, which include large-scale 4D egocentric dataset of human-object interactions. We leverage its 3D point cloud observations and corresponding MANO hand poses. To integrate this data, we process the MANO model parameters to generate our structural action representation $\mathbf{A} \in \mathbb{R}^{D_a \times T}$ and tag each joint according to our Embodied Joint Codebook (same for Robot and Simulation). For \textbf{robot demonstrations}, we incorporate 3D dexterous manipulation data from existing robot datasets, including Fourier ActionNet~\cite{fourier2025actionnet} and DexCap~\cite{wang2024dexcap}, which provide diverse 3D interaction trajectories. For the \textbf{simulation domains}, we collect expert demonstrations using policies trained via reinforcement learning (RL), such as VRL3-trained policies~\cite{rajeswaran2017learning} for Adroit~\cite{rajeswaran2017learning} and PPO-trained policies~\cite{schulman2017proximal} for DexArt~\cite{bao2023dexart} and Bi-DexHands~\cite{chen2023bi}. We filter these datasets to retain only successful trajectories.

\noindent\textbf{Implementation Details.}
We pre-train our model using the AdamW optimizer~\cite{loshchilov2017decoupled} with hyperparameters $\beta=(0.9,0.999)$, $\epsilon=1\times10^{-8}$ and weight decay $0.01$. We use a peak learning rate of $1\times10^{-4}$, scheduled with a linear warmup of $10{,}000$ steps followed by a cosine decay down to $1\times10^{-6}$. After the pre-training phase, the model is fine-tuned for each downstream simulation task using a small set of in-domain demonstrations with a fine-tuning learning rate of $1\times10^{-5}$. During inference, we generate the action chunk by solving the flow matching ODE. We employ Euler integration with a fixed step size of $10$ to map the noise vector to a clean action.

\vspace{-2pt}
\subsection{Comparison with Baselines}
\vspace{-2pt}

\noindent\textbf{Benchmarks and Simulation Environments.}
We evaluate our method by fine-tuning the pre-trained model on three challenging simulation benchmarks, totaling 11 tasks. Adroit~\cite{rajeswaran2017learning} is a  suite of difficult dexterous manipulation tasks using the Shadow Hand. DexArt~\cite{bao2023dexart} focuses on manipulating complex articulated objects. Bi-DexHands~\cite{chen2023bi} features a bimanual dexterous manipulation benchmark. These benchmarks are built upon high-fidelity physics simulators, such as MuJoCo~\cite{todorov2012mujoco} and IsaacGym~\cite{makoviychuk2021isaac}, which provide the realistic contact dynamics necessary for evaluating dexterous control.

\noindent\textbf{Baseline Methods.}
We compare our method with several strong baselines, as summarized in Table~\ref{tab:sim}. These include Diffusion Policy (DP)~\cite{chi2023diffusion} and HPT~\cite{wang2024scaling}, which are state-of-the-art policies that primarily rely on 2D images; UniAct~\cite{zheng2025universal}, a framework based on universal action tokens; and recent 3D-based approaches such as 3D Diffusion Policy (3DDP)~\cite{ze20243d} and 3D ManiFlow Policy~\cite{yan2025maniflow}, which directly use 3D point cloud inputs similar to ours. For all baselines, we use their official, publicly available implementations to ensure a fair comparison. We present the results of 3D realizations of the 2D baselines in the Appendix.

\begin{table}[t]
\centering
\small
\resizebox{0.46\textwidth}{!}{
\begin{tabular}{c|cc|c}
\toprule
Token Dim. & Prams. (M) & 1-NFE FLOPs (G) & Success \\
\midrule
16 & 4.71 & 0.66 & \dd{0.66}{0.02} \\
32 & 8.65 & 0.77 & \ddbfgreen{0.71}{0.05} \\
64 & 19.36 & 0.99 & \ddbfgreen{0.71}{0.04} \\
128 & 52.08 & 1.65 & \ddbfgreen{0.71}{0.03} \\
256 & 162.74 & 3.63 & \dd{0.70}{0.04} \\
\bottomrule
\end{tabular}
}
\vspace{-8pt}
\caption{Ablation on temporal compression and token dimension. We vary the token dimension ($d_{feat}$), which is the embedding size after compressing the temporal trajectory ($T$) of each joint. }
\vspace{-16pt}
\label{tab:time_compre}
\end{table}


\noindent\textbf{Key Findings.}
The results across 11 tasks are shown in
Table~\ref{tab:sim}. Our method consistently outperforms all 2D
and 3D baselines across the 11 simulation tasks. 
This superior performance is achieved with a remarkably efficient model.
As shown in Table~\ref{tab:sim}, our \mymethod{} has only \textbf{19.36M} parameters excluding the T5 tokenizer,
which is an order of magnitude smaller than 2D baselines, and significantly more compact than other 3D-based methods.
This highlights that our proposed structural-centric representation
 is not only effective for dexterous 3D manipulation but also parameter-efficient.

\vspace{-2pt}
\subsection{Ablation Studies}
\vspace{-2pt}

\noindent\textbf{Temporal Dimension Compression.} In Table~\ref{tab:time_compre}, we ablate the \textbf{Token Dim.} $d_{feat}$. This dimension impacts both parameters and FLOPs. We observe that performance is robust to compression, only dropping slightly when the dimension is reduced to 16. 
This slight degradation suggests that at extremely high compression
ratios, critical information begins to be
lost. Nonetheless, the policy's strong resilience to compression
down to 32 dimensions indicates that the temporal trajectories contain significant redundancy that can be compressed without substantial loss of performance.

\noindent\textbf{Pre-training Data
Composition.} Table~\ref{tab:data_scale} analyzes the impact of different
pre-training data combinations. While the full mixture yields the best result,
we find it very interesting that pre-training on Human data alone outperforms
pre-training on Robot data. This suggests our Embodied Joint Codebook
successfully translates functional human motion to robotic control.
Conversely, pre-training solely on simulation data outperforms
both Human-only and Robot-only pre-training. 
This is expected, as the simulation data is directly aligned with the
downstream fine-tuning tasks in terms of observation modality, 
action space, and environment dynamics.

\noindent\textbf{Model Component Ablation.}
In Table~\ref{tab:ablation}, we ablate key architectural components. While
removing observation features (global or local) or the causal mask leads to a
moderate performance drop, we observe a more significant degradation
when reverting to a conventional temporal-centric representation (the last row).
 This result validates our core hypothesis 
that the structural-centric formulation is a more effective 
paradigm for these high-DoF manipulation tasks.
Besides, removing the joint embedding causes
catastrophic failure. This is expected: our action sequence $\mathbf{A} \in
\mathbb{R}^{D_a \times T}$ is inherently unordered. Without the Embodied Joint
Codebook, the Transformer has no mechanism to determine which trajectory
corresponds to which physical joint, making the learning task impossible.

\begin{table}[t]
\centering
\small
\resizebox{0.45\textwidth}{!}{
\begin{tabular}{ccc|c|c}
\toprule
Human & Robot & Simulation & Scale & Success \\
\midrule
\yes & \yes & \yes & 100\% & \ddbfgreen{0.71}{0.04} \\
\yes & \no & \no & 100\% & \dd{0.68}{0.04} \\
\no & \yes & \no & 100\% & \dd{0.66}{0.05} \\
\no & \no & \yes & 100\% & \dd{0.70}{0.03} \\
\yes & \yes & \yes & 10\% & \dd{0.68}{0.03} \\
\bottomrule
\end{tabular}
}
\vspace{-8pt}
\caption{Pre-training data ablation. We report the average success on simulation tasks after fine-tuning models pre-trained on different combinations of Human, Robot, and Simulation datasets.}
\vspace{-16pt}
\label{tab:data_scale}
\end{table}

\begin{figure}[t]
  \centering
  \includegraphics[width=0.99\linewidth]{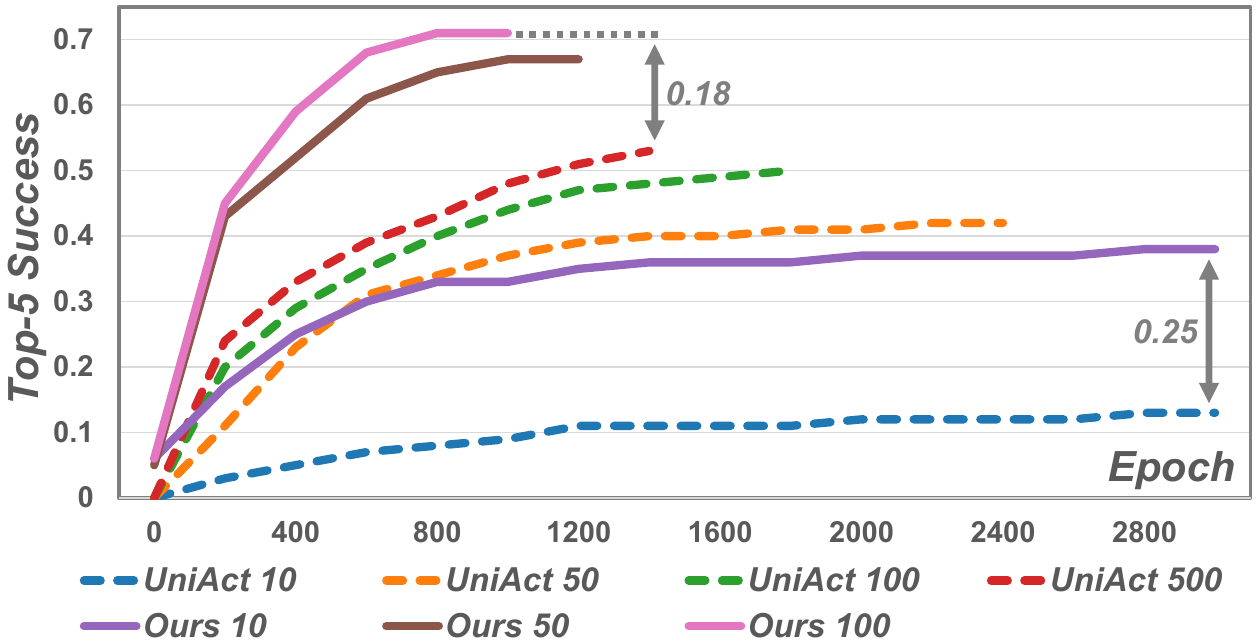}
  \vspace{-8pt}
  \caption{Few-shot adaptation efficiency. We plot the average
    success rate versus training epochs for our method and the
    UniAct~\cite{zheng2025universal} baseline, evaluated in
    few-shot settings using varying numbers of in-domain
    demonstrations.}
  \vspace{-12pt}
  \label{fig:finetune_efficiency}
\end{figure}

\noindent\textbf{Few-shot Adaptation Efficiency.}
To evaluate the quality of the learned prior, we compare its
few-shot adaptation efficiency against UniAct~\cite{zheng2025universal} in
Figure~\ref{fig:finetune_efficiency}. The results clearly show
that our method not only achieves a higher final success rate
across all settings but also learns significantly faster,
especially in low-data, few-shot regimes.

\begin{figure}[t]
  \centering
  \includegraphics[width=0.99\linewidth]{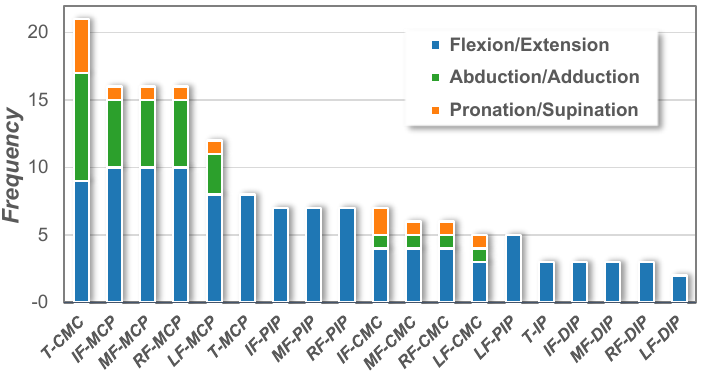}
  \vspace{-8pt}
  \caption{Frequency analysis of joint types in our Embodied Joint Codebook, derived from a survey of 10 common dexterous hands.
  }
  \vspace{-16pt}
  \label{fig:embodied_codebook}
\end{figure}

\begin{figure}[t]
  \centering
  \includegraphics[width=0.99\linewidth]{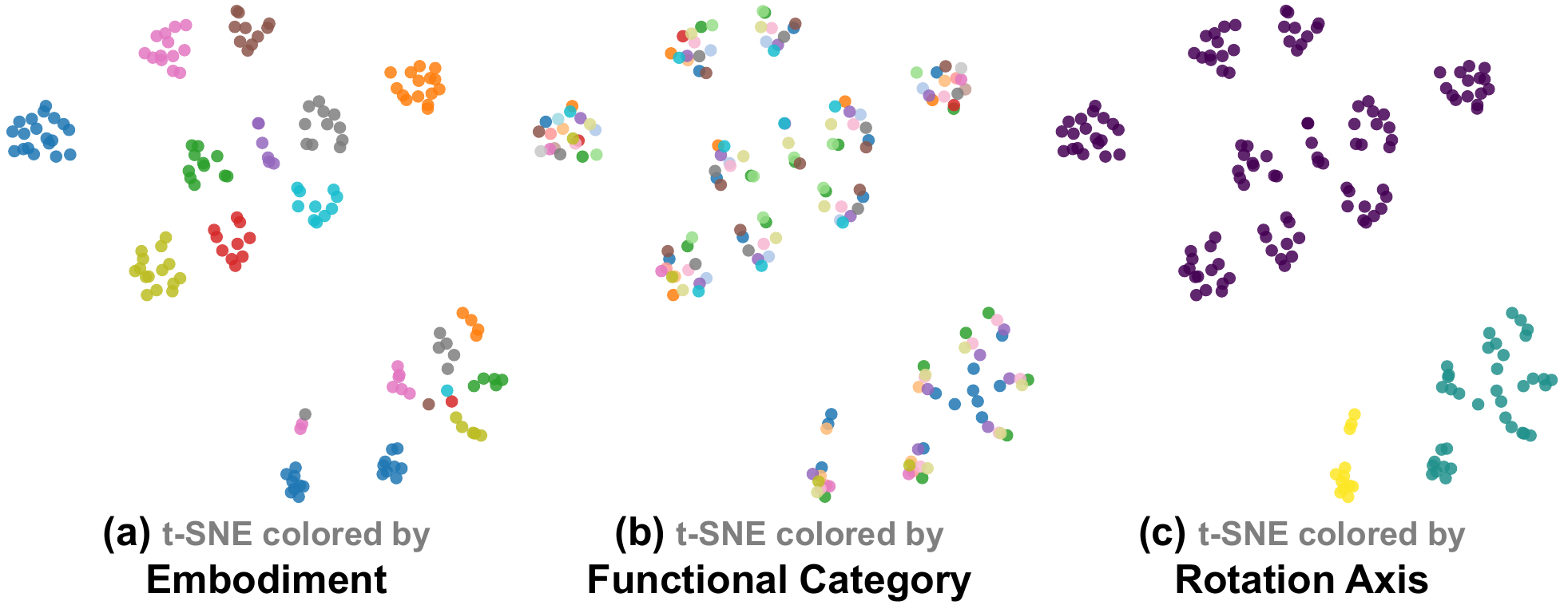}
  \vspace{-8pt}
  \caption{T-SNE visualization of the learned Embodied Joint Codebook embeddings.
    These embeddings, derived from 10 dexterous manipulators, are colored by \textbf{(a)} Embodiment ID, \textbf{(b)} Functional Category, and \textbf{(c)} Rotation Axis.}
  \vspace{-16pt}
  \label{fig:tsne}
\end{figure}

\noindent\textbf{Embodied Joint Codebook Analysis.} 
We further dissect the Embodied Joint Codebook in Table~\ref{tab:embodied_joint_codebook_abla}.
While removing the Embodiment ID or Rotation Axis leads to a performance drop, ablating the Functional Category causes a catastrophic failure.
This indicates that functional correspondence is the most critical factor for the policy to bridge the embodiment gap.
To understand why this works, we analyze 10 common dexterous manipulators to inform our codebook design, including the Shadow Dexterous Hand, Allegro Hand, Inspire Robots Dexterous Hand, \textit{etc}. In Figure~\ref{fig:embodied_codebook}, we plot the frequency of each joint type from our codebook across these hands. We find that Flexion/Extension joints for the MCP, CMC and PIP joints are the most frequent. This suggests these joints represent a core functional set that is essential for dexterous manipulation and should be prioritized in hardware design. 
We also visualize the learned codebook embedding space from
these hands using t-SNE in Figure~\ref{fig:tsne}. The
visualization illustrates clear, dominant clustering by
Embodiment (a) and Rotation Axis (c).
Functional Category (b) clusters are less distinct, as the model learns a highly specific embedding for each joint to resolve structural ambiguity.
This suggests that generalization stems not from embedding similarity, but from the codebook's compositional structure, which allows the model to learn transferable functional patterns.

\begin{table}[t]
\centering
\small
\resizebox{0.43\textwidth}{!}{
\begin{tabular}{l|c}
\toprule
Model Variant & Success \\
\midrule
\mymethod{} & \ddbfgreen{0.71}{0.04} \\
\mymethod{} w.o. global point cloud token & \dd{0.68}{0.05} \\
\mymethod{} w.o. local point cloud token & \dd{0.69}{0.03} \\
\mymethod{} w.o. causal mask & \dd{0.68}{0.04} \\
\mymethod{} w.o. joint embedding & \dd{0.01}{0.01} \\
\mymethod{} w. temporal-centric action & \dd{0.64}{0.05} \\
\bottomrule
\end{tabular}
}
\vspace{-8pt}
\caption{Model component ablation. We analyze the impact of removing key architectural components on the average success rate.}
\vspace{-12pt}
\label{tab:ablation}
\end{table}

\vspace{-2pt}
\subsection{Real-World Experiments}
\vspace{-2pt}

\begin{figure}[t]
  \centering
  \includegraphics[width=0.99\linewidth]{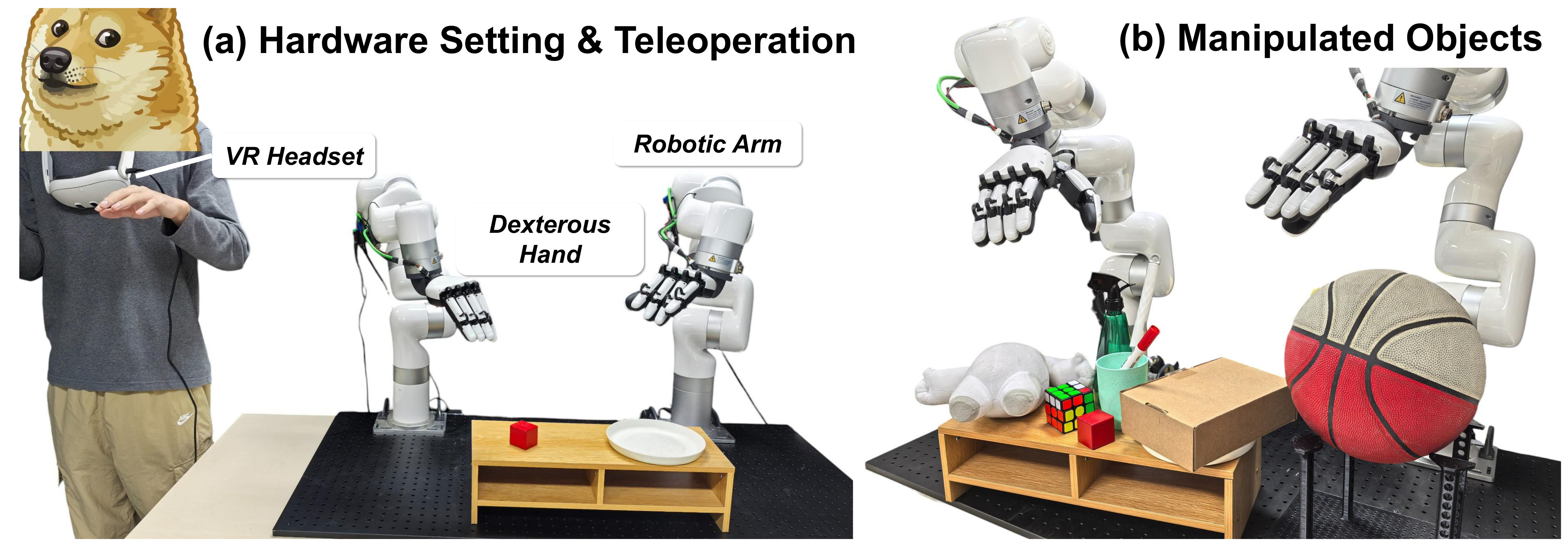}
  \vspace{-8pt}
  \caption{Real-world experimental setup. 
\textbf{(a)} Our bimanual hardware setup. We collect demonstration data using a VR headset for teleoperation. 
\textbf{(b)} The set of diverse objects used in our real-world manipulation, requiring both precision and bimanual coordination.}
  \vspace{-16pt}
  \label{fig:real_setup}
\end{figure}

\noindent\textbf{Hardware and Teleoperation Setup.}
Our real-world bimanual system, shown in Figure~\ref{fig:real_setup} (a), is comprised of two 7-DoF xArm robotic arms, each equipped with a 12-DoF xHand dexterous hand. We use a single L515 LiDAR camera mounted above the workspace for 3D scene perception, providing point cloud observations. The set of manipulated objects, seen in Figure~\ref{fig:real_setup} (b), includes a pen with a cap, a Baymax toy, a cardboard box, a plastic plate, a cup with a brush, and a basketball.

\begin{table}[t]
\centering
\small
\resizebox{0.45\textwidth}{!}{
\begin{tabular}{ccc|c}
\toprule
Embodiment & Function & Rotation & Success (11 tasks) \\
\midrule
\yes & \yes & \yes & \ddbfgreen{0.71}{0.04} \\
\no & \yes & \yes & \dd{0.57}{0.05} \\
\yes & \no & \yes & \dd{0.02}{0.02} \\
\yes & \yes & \no & \dd{0.62}{0.07} \\
\bottomrule
\end{tabular}
}
\vspace{-8pt}
\caption{Ablation study on the components of the Embodied Joint Codebook.
    We report the average success rate after fine-tuning models with different codebook components ablated.}
\vspace{-12pt}
\label{tab:embodied_joint_codebook_abla}
\end{table}

\noindent\textbf{Data Collection via Teleoperation.}
We collect demonstration data using a Meta Quest 3 VR headset, which captures the operator's hand and finger motions. To transfer the human motion to the robot's dissimilar kinematics, we employ a real-time retargeting strategy from AnyTele~\cite{qin2023anyteleop}. 
Specifically, our teleoperation system reads the 3D poses of the operator's hand links from the VR controller. It then optimizes the xHand's joint angles by minimizing an objective function defined as the difference between the wrist-to-fingertip vectors on the human hand and the corresponding vectors on the robot hand. This method allows for intuitive control and the collection of dexterous demonstration data.

\begin{figure*}[t]
  \centering
  \includegraphics[width=0.96\linewidth]{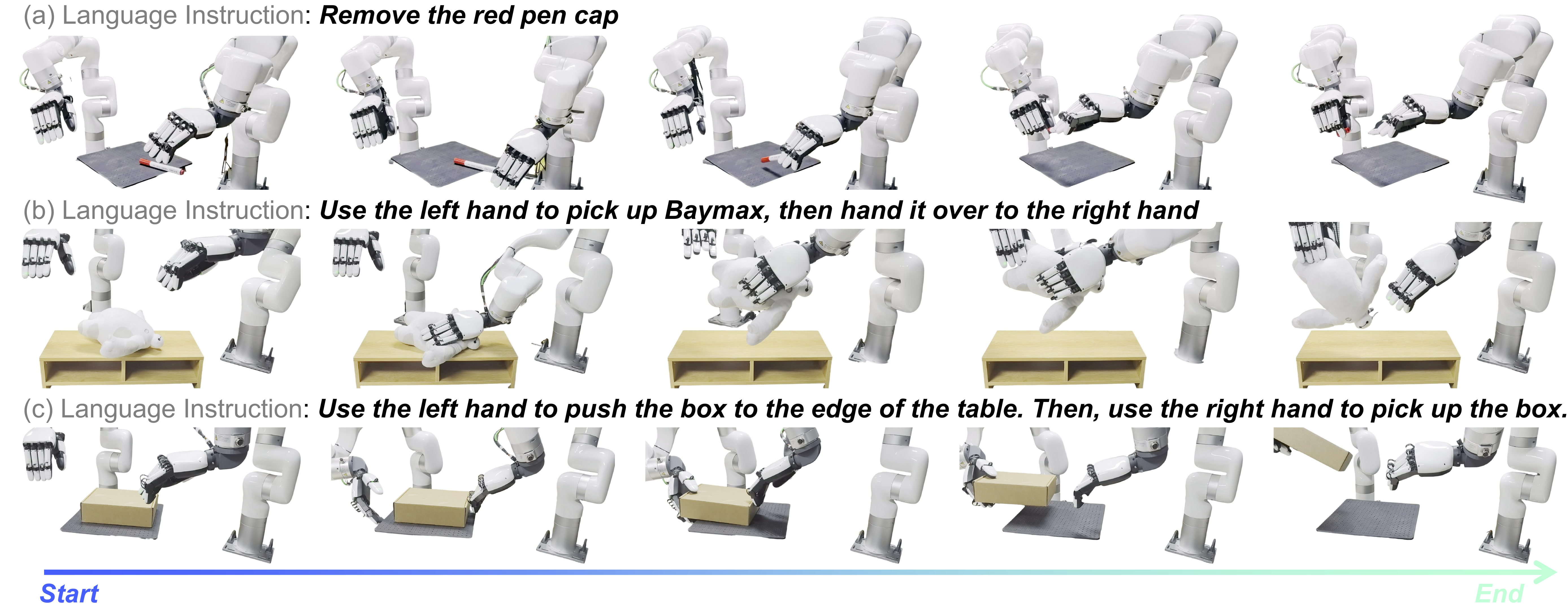}
  \vspace{-16pt}
  \caption{Qualitative rollouts of our policy executing complex bimanual tasks in the real world.}
  \vspace{-16pt}
  \label{fig:real_traj}
\end{figure*}

\noindent\textbf{Task Suite and Experimental Setup.}
We evaluate performance on 6 challenging tasks, which require both single-arm precision and complex bimanual coordination.
\begin{itemize}
    \setlength\itemsep{-0.2em}
    \item \textbf{Remove the pen cap:} A bimanual task where one hand must stabilize the pen body while the other grasps and removes the cap (Figure~\ref{fig:real_traj} (a)).
    \item \textbf{Hand over Baymax:} A coordinated bimanual task where the left hand picks up the toy and passes it to the right hand (Figure~\ref{fig:real_traj} (b)).
    \item \textbf{Push then grab box:} A long-horizon, multi-stage bimanual task. The left hand first pushes the box to the edge of a platform, after which the right hand grasps it from the table (Figure~\ref{fig:real_traj} (c)).
    \item \textbf{Place block in plate:} A single-arm task testing grasping and placement of a block into a target container.
    \item \textbf{Brush the cup:} A contact-rich bimanual task where one hand holds the cup, and the other uses a brush to perform a cleaning motion inside it.
    \item \textbf{Grasp basketball:} A bimanual task requiring the two hands to cooperatively form a stable, large-volume grasp on the ball.
\end{itemize}
\noindent Qualitative rollouts for the remaining tasks are included
in the Appendix.

\noindent\textbf{Experimental Setup.}
For each of the 6 tasks, we collect a dataset of 50 successful demonstrations. We compare three methods:
HPT~\cite{wang2024scaling} uses its publicly available pre-trained weights. We train task-specific input stems and heads for each task.
3DDP~\cite{ze20243d} is trained from scratch for each task.
\mymethod{} uses the pre-trained model.
We then fine-tune this model on the demonstrations, evaluating its ability to adapt its structural priors to the real-world domain.

\noindent\textbf{Real-World Results.}
The quantitative results are summarized in Table~\ref{tab:real_results}. Our method achieves higher success rates than all baselines across 6 tasks. This suggests that our structural action representation pre-trained on diverse data, provides a powerful prior for learning complex bimanual coordination. Figure~\ref{fig:real_traj} shows qualitative rollouts of our policy executing three of the bimanual tasks.

\vspace{-2pt}
\subsection{Failure Case Analysis and Limitations}
\vspace{-2pt}

In addition to the quantitative ablations, we analyze failure modes which highlight the limitations of our current approach and provide avenues for future work. In complex, contact-rich bimanual tasks, one hand may severely occlude the other hand or the target object. A single, fixed camera provides no mechanism to resolve this ambiguity, starving the policy of the critical geometric information needed for precise coordination. We also observe large kinematic or contact-geometry mismatches between demonstrations and the execution platform produce action-assignment errors. These failures require explicit dynamics/force constraints or tactile closed-loop correction to resolve. We provide qualitative visualizations
of these failure modes in the Appendix.

\begin{table}[t]
\centering
\small
\resizebox{0.48\textwidth}{!}{
\begin{tabular}{l|ccc}
\toprule
  & \multicolumn{3}{c}{Success Rate} \\
 Task Description & HPT~\cite{wang2024scaling} & 3DDP~\cite{ze20243d} & \textbf{\mymethod{} (Ours)} \\
\midrule
\textit{Remove the pen cap} & 0.10 & 0.25 & {\cellcolor{tablecolor2}$\mathbf{0.30}$} \\
\textit{Hand over Baymax} & 0.50 & 0.75 & {\cellcolor{tablecolor2}$\mathbf{0.85}$} \\
\textit{Push then grab box} & 0.05 & 0.15 & {\cellcolor{tablecolor2}$\mathbf{0.35}$} \\
\textit{Place block in plate} & 0.60 & 0.85 & {\cellcolor{tablecolor2}$\mathbf{0.90}$} \\
\textit{Brush the cup} & 0.10 & 0.30 & {\cellcolor{tablecolor2}$\mathbf{0.45}$} \\
\textit{Grasp basketball} & 0.65 & 0.80 & {\cellcolor{tablecolor2}$\mathbf{0.95}$} \\
\bottomrule
\end{tabular}
}
\vspace{-8pt}
\caption{Quantitative results on 6 real-world bimanual manipulation tasks. We compare our model against baselines using in-domain demonstrations.}
\vspace{-16pt}
\label{tab:real_results}
\end{table}

\vspace{-6pt}
\section{Conclusion}
\vspace{-4pt}

In this work, we introduce a fundamental shift in action representation for policy learning, challenging the conventional temporal-centric paradigm.
We propose a \textit{structural-centric} representation, which models an action chunk as a variable-length sequence of joint-wise trajectories.
We demonstrate that this structural formulation provides a
powerful framework for tackling cross-embodiment skill
transfer in high-DoF dexterous hands.
Through comprehensive evaluation across both simulation and real-world bimanual manipulation tasks, our method consistently outperforms all baselines.
Our results highlight that a structural-centric prior enables temporal compression and allows the policy to bridge the morphological gap between demonstrators and robots.
In future research, this structural action representation will be extended beyond imitation learning, investigating its potential as an expressive policy class for reinforcement learning, where it could offer a structured exploration space for complex high-DoF agents.

\section{Acknowledgments}

This work was supported by the National Natural Science Foundation of China under Contract 62472141, the Natural Science Foundation of Anhui Province under Contract 2508085Y040, and the Youth Innovation Promotion Association CAS. It was also supported by the GPU cluster built by MCC Lab of Information Science and Technology Institution, USTC and the Supercomputing Center of USTC.

{
    \small
    \bibliographystyle{ieeenat_fullname}
    \bibliography{main}
}


\end{document}